\documentclass[letterpaper, 10 pt, conference]{ieeeconf}  

\IEEEoverridecommandlockouts                              

\let\labelindent\relax

\usepackage{amsmath,amsfonts,bm}

\def\eqref#1{equation~\ref{#1}}

\def\1{\bm{1}}

\DeclareMathAlphabet{\mathsfit}{\encodingdefault}{\sfdefault}{m}{sl}
\SetMathAlphabet{\mathsfit}{bold}{\encodingdefault}{\sfdefault}{bx}{n}

\usepackage{hyperref}
\usepackage{url}
\usepackage{enumitem}
\usepackage{subcaption}     
\usepackage{booktabs}       
\usepackage{cite}
\usepackage{graphicx}
\usepackage{xcolor}
\usepackage{pifont}
\usepackage{caption}
\newcommand{\cmark}{\ding{51}}%
\newcommand{\xmark}{\ding{55}}%

\newcommand{\ie}{\textit{i}.\textit{e}.}

\newcommand{\etc}{\textit{etc}.}

\newcommand{\greencheck}{{\color{green}\cmark}}
\newcommand{\redcross}{{\color{red}\xmark}}

\title{\LARGE \bf
HexPlane Representation for 3D Semantic Scene Understanding
}

\author{Zeren Chen$^{1*}$, Yuenan Hou$^{2*}$$\dagger$, Yulin Chen$^{2}$, Li Liu$^{3}$, Xiao Sun$^{2}$, Lu Sheng$^{1}$$\dagger$
\thanks{*: Equal contributions. $\dagger$: Corresponding authors.}
\thanks{$^{1}$Zeren Chen and Lu Sheng are with School of Software, Beihang University
        {\tt\small \{czr1604, lsheng\}@buaa.edu.cn}}%
\thanks{$^{2}$Yuenan Hou, Yulin Chen and Xiao Sun are with Shanghai AI Laboratory
        {\tt\small \{houyuenan, chenyulin, sunxiao\}@pjlab.org.cn}}%
\thanks{$^{3}$Li Liu is with the College of Electronic Science and Technology, National University of Defense Technology (NUDT)
        {\tt\small liuli\_nudt@nudt.edu.cn}}
}

\begin{document}

\maketitle
\thispagestyle{empty}
\pagestyle{empty}

\def\algorithmname{HexNet3D}

\begin{abstract}	
In this paper, we introduce the HexPlane representation for 3D semantic scene understanding. Specifically, we first design the View Projection Module (VPM) to project the 3D point cloud into six planes to maximally retain the original spatial information. Features of six planes are extracted by the 2D encoder and sent to the HexPlane Association Module (HAM) to adaptively fuse the most informative information for each point. The fused point features are further fed to the task head to yield the ultimate predictions. Compared to the popular point and voxel representation, the HexPlane representation is efficient and can utilize highly optimized 2D operations to process sparse and unordered 3D point clouds. It can also leverage off-the-shelf 2D models, network weights, and training recipes to achieve accurate scene understanding in 3D space. On ScanNet and SemanticKITTI benchmarks, our algorithm, dubbed \algorithmname, achieves competitive performance with previous algorithms. In particular, on the ScanNet 3D segmentation task, our method obtains \textbf{77.0} mIoU on the validation set, surpassing Point Transformer V2 by \textbf{1.6} mIoU. We also observe encouraging results in indoor 3D detection tasks. Note that our method can be seamlessly integrated into existing voxel-based, point-based, and range-based approaches and brings considerable gains without bells and whistles. The codes will be available upon publication.
\end{abstract}

\section{INTRODUCTION}
\label{sec:intro}

3D semantic scene understanding aims to provide comprehensive and fine-grained categorical information about the surrounding environment, which is pivotal to the perception, planning, and control of autonomous vehicles and embodied systems. A plethora of excellent research works have emerged in this field, such as the PointNet series~\cite{pointnet,pointnet++}, SPVCNN~\cite{spvcnn}, Cylinder3D~\cite{cylinder3d}, the Point Transformer series~\cite{ptv1,ptv2}, \etc. Point and voxel representations are favored by these powerful models. However, the point representation requires the processing of a massive amount of raw points while the voxel representation is hindered by the information loss caused by the voxelization operation and needs specifically designed sparse convolution to handle. It is natural to wonder if we can design a representation that can enjoy the benefits of both worlds.

\begin{table}[ht!]
    \centering
    \resizebox{0.6\linewidth}{!}{
    \begin{tabular}{l|cccccc}
        \toprule
         Representations &  \rotatebox[origin=l]{90}{Efficient} & \rotatebox[origin=l]{90}{Dense operation} & \rotatebox[origin=l]{90}{Handle occlusion} & \rotatebox[origin=l]{90}{2D knowledge}  \\ \midrule
         Point &    \redcross         &    \redcross       &     \greencheck      &   \redcross       \\
         Voxel &     \greencheck     &    \redcross    &   \greencheck       &    \redcross   \\
         Range &     \greencheck     &   \greencheck       &    \redcross       & \greencheck       \\ \midrule 
         HexPlane & \greencheck  &  \greencheck &  \greencheck  &  \greencheck     \\ \bottomrule
    \end{tabular}}
    \caption{Comparison with different representations. Here ``Efficient'' means being free from handling all raw point clouds, and ``Dense operation'' denotes models that can utilize dense and highly optimized 2D operations. ``Handle occlusion'' denotes models that can exhibit good performance in scenes with severe occlusions. ``2D knowledge'' denotes borrowing knowledge from off-the-shelf 2D pre-trained models.}
    \label{tab:intro}
\end{table}

To this end, we introduce the HexPlane representation for the 3D scene understanding tasks. As the name implies, we first design the View Projection Module (VPM) to project the 3D point cloud into six planes to maintain the original spatial information to the maximum extent. These six planes are empirically defined as the XZ plane (front, back), YZ plane (left, right), XY plane (top view) and the cylindrical plane as the combination of these planes can maximally retain the information and yield the best performance. Then, the features of six planes are extracted by the 2D encoder and fed to the HexPlane Association Module (HAM). For each point, HAM can adaptively choose the most informative information of each plane to fuse. Thereafter, the fused hexplane features, accompanied with the pointwise features, are sent to the task head to yield the predictions. 

As shown in Table.~\ref{tab:intro}, compared to the point and voxel representations, the HexPlane representation is much more efficient and can utilize the highly optimized 2D operations to process the point clouds. It can also leverage off-the-shelf 2D models, network weights as well as the training recipes to fulfill precise 3D scene understanding. The HexPlane representation can inject the abundent 2D knowledge into the 3D world, easily utilize the rich knowledge of the off-the-shelf 2D models, and relieve the data scarcity problem in the 3D field. Note that the designed HexPlane representation can be seamlessly integrated into the point/voxel-based models and bring considerable gains with little computation overhead. We perform extensive experiments on ScanNet~\cite{scannet} and SemanticKITTI~\cite{semantickitti} benchmarks and our HexNet3D achieves competitive against previous competitive networks in 3D semantic segmentation tasks. Encouraging results are also observed in indoor 3D detection tasks. We hope our HexNet3D can serve as the footstone of the 3D perception algorithms.

To sum up, the contributions of this paper are listed as follows:

\begin{itemize}[leftmargin=*]
\item We introduce the HexPlane representation for 3D scene understanding tasks. It is efficient, effective and can leverage the rich knowledge in 2D field to help accomplish 3D tasks.

\item We design the View Projection Module (VPM) and HexPlane Association Module (HAM) to process the point features and achieve adaptive fusion of HexPlane features, respectively.

\item We perform extensive experiments on ScanNet and SemanticKITTI benchmarks. Notably, our method achieves \textbf{77.0} mIoU on the validation set of ScanNet semantic segmentation task, outperforming the competitive Point Transformer V2 by \textbf{1.6} mIoU.

\end{itemize}

\section{RELATED WORK}
\label{sec:relatedwork}
In this section, we have a brief review of relevant field, including LiDAR-based scene understanding, multi-modal LiDAR segmentation and knowledge transfer in 3D scene understanding tasks.

\subsection{LiDAR-based scene understanding}
LiDAR-based scene understanding has recently witnessed rapid development and a large quantity of algorithms have been put forward~\cite{survey_3d}. According to the type of the input signal, these methods can be categorized into point-, voxel- and range-based approaches. For point-based models, KPConv~\cite{kpconv} devises the kernel point convolution to directly process the input points. For voxel-based methods, MinkovskiNet~\cite{choy20194d} designs the UNet-based model to process the voxelized point cloud sequences. SPVCNN~\cite{spvcnn} attaches the point branch to the voxel branch of MinkovskiNet to supplement fine-grained pointwise information. Cylinder3D~\cite{cylinder3d} proposes to replace the cubic partition of point cloud with cylindrical partition to better handle the varying density of point cloud. For the range-based models, RangeFormer~\cite{rangeformer} and RangeViT~\cite{rangevit} capitalize on the powerful transformer-based architecture to perform LiDAR segmentation given the projected range images. The proposed HexPlane representation is efficient and can use highly-optimized 2D operations, and benefit from the rich knowledge in the 2D field. Also, our method is complementary to the above-mentioned LiDAR segmentation models and can be flexibly plugged into existing framework without much efforts.
LiDAR-based object detection has also witnessed significant strides in recent years~\cite{fcaf3d,3d-mpa,3detr,hgnet,votenet,gsdn}. FCAF3D~\cite{fcaf3d} is a competitive fully convolutional anchor-free 3D detector and has achieved impressive performance in indoor scenarios. 3DETR~\cite{3detr} designs a transformer-based network for end-to-end 3D object detection. 

\subsection{Multi-modal LiDAR segmentation} 
Although point cloud can provide precise 3D positions of objects, the lack of color and texture makes the LiDAR-based model fail to utilize the valuable information from the surrounding environment. Images are natural complement to the point cloud and the multi-modal solution is gaining surging attention~\cite{mm_survey,bev_survey,occmamba}. UniSeg~\cite{uniseg} capitalizes on the complementary knowledge of point, voxel, range images and RGB images to develop more comprehensive and robust perception system. MSeg3D~\cite{mseg3d} designs effective strategies to achieve better multi-modal fusion from the perspective of modality heterogeneity, FOV (Field Of View) differences and data augmentation. TASeg~\cite{taseg} designs a novel and efficient multi-frame concatenation and fusion framework for utilizing the valuable spatial and temporal information in point cloud and image sequences. The Point Transformer series open the possibility of applying transformer-based architectures to indoor RGB-D scene understanding~\cite{ptv1,ptv2}.

\subsection{Knowledge transfer in 3D tasks} 
Knowledge transfer is originally designed to transfer the knowledge from large cumbersome teacher model to compact student model. PVKD~\cite{pvkd} makes the first attempt to apply knowledge distillation to the 3D semantic segmentation task by designing the 3D-aware teacher-student learning paradigm. Since the collection and annotation of point clouds is tedious and expensive, various attempts~\cite{seal,clip2scene,pointclip,slidr} have been made to transfer the rich knowledge from other modalities to the point cloud domain, \emph{e.g.}, image and text. PointCLIP~\cite{pointclip} is the pioneering work that conveys the multi-modal knowledge from CLIP to the point cloud understanding tasks. SLiDR~\cite{slidr} resorts to the contrastive objective to inject the rich 2D knowledge into the 3D world. CLIP2Scene~\cite{clip2scene} makes the first attempt to apply CLIP to LiDAR-based semantic segmentation. PointLLM~\cite{pointllm} explores the way to utilize the valuable textual information extracted from large language models to guide the learning of point cloud models. As opposed to the aforementioned algorithms, we utilize the 2D models, pre-trained weights as well as the training recipe for 3D scene understanding.


\section{METHOD}
\label{sec:method}

\begin{figure*}[t]
\begin{center}
\setlength{\fboxrule}{0pt}
\fbox{\includegraphics[width=1.0\textwidth]{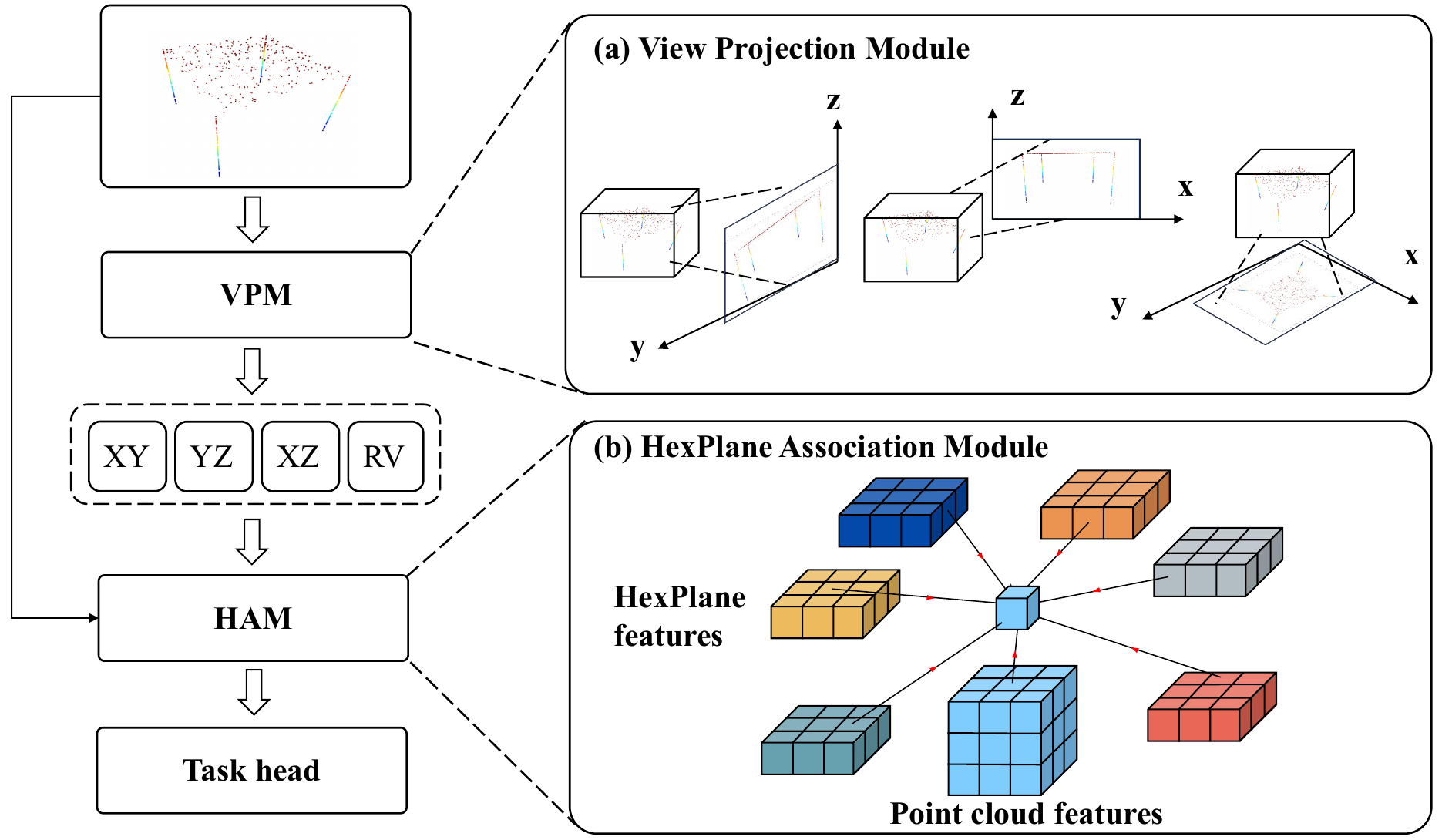}}
\end{center}
\caption{Framework overview of our \algorithmname~framework, which is comprised of (a) View Projection Module (VPM) and (b) HexPlane Association Module (HAM). Given the input point cloud, we first leverage VPM to project the point into six planes. The HexPlane features are produced by the 2D encoder and then fed to HAM to adaptively fuse the most informative information for each point. The fused point features are eventually sent to the task head, thus yielding the predictions.}
\label{fig:overview}
\end{figure*}

In this section, we first briefly revisit the design of RangeFormer, a transformer-based framework that projects point cloud features onto a range view and then processes them with a 2D backbone, i.e., SegFormer. We then describe our proposed approach that mitigates the inherent occlusion issues of the range view by introducing a novel HexPlane representation for 3D scene perception. Eventually, the overall training objective is presented. The framework overview of our \algorithmname~is shown in Fig.~\ref{fig:overview}.

\subsection{Revisiting RangeFormer}

RangeFormer~\cite{rangeformer} projects the 3D point cloud into a 2D range image by rasterizing the raw point cloud data. In its original design, the point cloud is first projected onto a cylindrical range view and then fed into a 2D backbone, i.e., SegFormer, for segmentation. Although this approach leverages efficient and well-established 2D operations, it suffers in highly dense scenes where occlusions lead to the loss of critical information. Increasing the resolution of the range view can alleviate this issue to a certain extent, but at the expense of significantly higher computational costs. Note that we merely adopt the backbone of RangeFormer.

\subsection{HexPlane Representation for 3D Segmentation}

Let $\mathbf{P} \in \mathbb{R}^{N \times 3}$ denote the input point cloud, where $N$ is the number of points. Directly processing the point cloud with 2D vision models is intractable due to the inherent sparsity of point cloud. 

\subsection{View Projection Module}

\begin{figure}[t]
\begin{center}
\setlength{\fboxrule}{0pt}
\fbox{\includegraphics[width=0.45\textwidth]{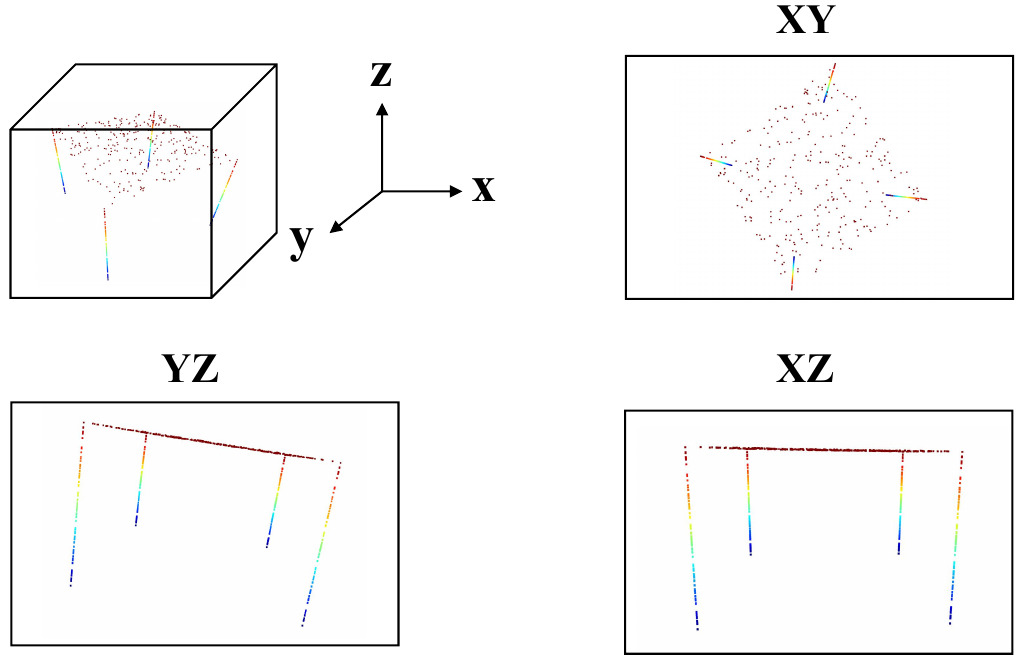}}
\end{center}
\caption{View projections of the input point cloud, including the XY plane, YZ plane and XZ plane. Different view projections are complementary to each other and provide valuable visual clues for accurate 3D semantic scene understanding.}
\label{fig:vpm}
\end{figure}

Therefore, we project the point cloud into six 2D planes, namely the XY, $\pm$XZ, $\pm$YZ, and the cylindrical range view, $V \in \mathbb{R}^{M \times H \times W \times D}$. The view projections of the point cloud are shown in Fig.~\ref{fig:vpm}. Since the XY, $\pm$XZ, $\pm$YZ projections are straightforward, we put more emphasis on the cylindrical view projection process.  

For a given LiDAR point cloud, we rasterize the points into a 2D cylindrical projection $\mathcal{R}(u,v)$ (\textit{a.k.a.}, range image) of size $H\times W$, where $H$ and $W$ are the height and width of the range image, respectively. The rasterization process for each point $p_n$ can be formulated as follows:
\begin{equation}
\label{eq:rv}
\begin{pmatrix}
\mathit{u}_n  \\
\mathit{v}_n
\end{pmatrix}
=
\begin{pmatrix}
\frac{1}{2}~[1-\arctan(p^y_n,p^x_n)\pi^{-1}]~W  \\
~[1-(\arcsin(p^z_n,{(p^d_n)}^{-1})+\phi^{\text{down}})\xi^{-1}]~H~
\end{pmatrix},
\end{equation} 
where $(u_n,v_n)$ denotes the grid coordinate of point $p_n$ in range image $\mathcal{R}(u,v)$; $p^d_n=\sqrt{(p^x_n)^2+(p^y_n)^2+(p^z_n)^2}$ is the depth between the point and LiDAR sensor (ego-vehicle); $\xi=|\phi^{\text{up}}|+|\phi^{\text{down}}|$ denotes the vertical field-of-views (FOVs) of the sensor and $\phi^{\text{up}}$ and $\phi^{\text{down}}$ are the inclination angles at the upward and downward directions, respectively. Note that $H$ is often predefined by the beam number of the LiDAR sensor, while $W$ can be set based on requirements.

The detailed view projection process is shown as follows:
For each 2D view $V_i$, we feed it to the 2D backbone (\ie, SegFormer), generating six 2D view features $F^I \in \mathbb{R}^{M \times H_f \times W_f \times C_f}$ where $H_f, W_f, C_f$ is the corresponding height, width and the number of channels of the feature map. Note that we fuse the multi-scale features output by SegFormer. For simplicity of notation, we omit this. The six projections together provide complementary viewpoints that help alleviate occluded details and minimize information loss.

\subsection{Injecting the 2D Knowledge into the 3D World}  

While the 2D backbone yields rich semantic features from each projected view, not all 3D points are reliably mapped to every plane. Thus, instead of directly performing perception solely on the projected planes, we instead inject the expressive and informative knowledge from off-the-shelf 2D backbone into the 3D point cloud. How to effectively fuse the 2D and 3D information is vital to the segmentation results. One straightforward way is to concatenate the 2D features with the original 3D point cloud features $F^P$ based on the point index. However, there are several shortcomings of this practice, e.g., not all points and view features can be aligned one-to-one.

\subsection{HexPlane Association Module}

To overcome this issue, we propose a dedicated HexPlane Association Module (HAM) based on cross attention mechanism. In this module, the 3D point cloud features $F^P$ extracted by a lightweight 3D CNN serve as queries $q$, while the 2D view features $F^I$ act as keys $k$ and values $v$:
$$
\bm{p}=\texttt{CrossAttn}(F^I,F^P,\phi)=\texttt{MHA}(\bm{q},\bm{k},\bm{v})
$$
$$
\bm{q}=f^q(F^P);\;\bm{k}=f^k(F^I);\;\bm{v}=f^v(F^I)
$$
Here, $f^q, f^k, f^v$ are corresponding projection heads. In addition, we also incorporate a position embedding $\phi$ computed from the difference between the spatial locations of the 3D query points and those of points projected on views. In cases where a point is exactly projected onto the view, the positional offset is zero, while if the projection is occluded, the offset is larger. We then use $\bm{p}$ for subsequent segmentation via a segmentation head. This strategy effectively injects 2D contextual knowledge into the 3D domain, leading to more robust segmentation performance.

\subsection{Training Objective}

As to the LiDAR segmentation tasks, the final training loss is a composite of the main segmentation loss and auxiliary losses applied to the intermediate outputs of the 2D branches. By supervising both the fused 3D predictions and the individual 2D view predictions, the network is encouraged to learn complementary representations that enhance overall segmentation accuracy. As for the LiDAR detection task, the ultimate loss objective comprises the bounding box regression loss and the category classification loss.

\section{EXPERIMENTS}
\label{sec:experiments}

In this section, we present the used benchmarks, evaluation metrics, implementation details, evaluation protocal as well as the quantitative results.

\subsection{Benchmarks \& Metrics} 

Following the practice of previous 3D scene understanding algorithms~\cite{ptv1,ptv2,fcaf3d,cenet,rangeformer}, we perform extensive experiments on ScanNetV2 (segmentation and detection)~\cite{scannet} and SemanticKITTI (segmentation)~\cite{semantickitti}. 
The ScanNetV2 contains 1, 513 large-scale 3D scans reconstructed from RGB-D frames. 1, 201 scanned scenes are used for training and 312 scenes are chosen for validation.
The point clouds are sampled from vertices of reconstructed meshes, and each sampled point is assigned a semantic label from 20 categories, \emph{e.g.}, wall, floor, cabinet, \emph{etc}. 
The SemanticKITTI is a large-scale autonomous driving benchmark. It is comprised of 22 point cloud sequences, where sequences 00 to 10, 08, and 11 to 21 are used for training, validation and testing, respectively. The total number of categories is 19. The 19 classes are chosen for training and evaluation after merging classes with distinct moving status and discarding classes with very few points.
For semantic segmentation tasks, we adopt the Intersection-over-Union (IoU), mean of Intersection-over-Union (mIoU), mean of class-wise accuracy (mAcc), and overall point-wise accuracy (OA) as the evaluation criterion. For the detection task, we follow FCAF3D~\cite{fcaf3d} and take mAP$_{25}$ and mAP$_{50}$ as the evaluation metric. The detailed definition of the used metrics are presented as follows.

We first use Intersection-over-Union (IoU) to measure the distance between the predictions and labels. The definition of IoU between prediction A and label B is given below: 
\begin{equation}
\mathbf{IoU}(A, B) = \frac{|A \bigcap B|}{|A \bigcup B|}.
\end{equation} 
If the IoU between prediction and its label is larger than a pre-defined value, the prediction is seen as True Positive (TP). Otherwise, it is treated as False Positive (FP). Then, Precision and Recall can be computed based on TP, FP, and FN: 
\begin{equation}
\mathbf{Precision} = \frac{TP}{TP + FP}, \mathbf{Recall} = \frac{TP}{TP + FN}. 
\end{equation}
Here, FN denotes False Negative. And AP is calculated using the interpolated precision values: 
\begin{equation}
\mathbf{AP} = \frac{1}{|R|}\sum_{r \in R}{p_{interp}(r)},
\end{equation}
where $R$ is the set of all recall positions, $p_{interp}(.)$ is the interpolation function, defined as: $p_{interp}(r) = \max_{r':r'\geq r}p(r')$, and the mean average precision (mAP) is the average of APs of different classes or difficulty levels.

\subsection{Implementation details} 
HexPlane builds on a transformer-based backbone (SegFormer) that extracts multi-scale features from multiple 2D view projections of the 3D point cloud. The input views are generated at resolutions of 256$\times$256 for XY plane and 64$\times$512 for another 5 planes, ensuring sufficient spatial granularity. Besides, we apply a lightweight 12-layer 3D CNN for extracting point features. Then a dedicated point fusion neck aggregates information from these diverse views to strengthen the point features, and separate segmentation heads are employed for both point-level and view-level predictions. Training is performed with a batch size of 8 per GPU over 150 epochs. We use the AdamW optimizer and OneCycle scheduler with an initial learning rate of 3.5e-4 and a weight decay of 0.01. Except for regular data augmentations like random flipping and rotating, we also apply a point mix augmentations following~\cite{rangeformer}. These design choices collectively address the challenges of sparsity and deformation inherent in range view representations. 

\subsection{Evaluation Protocal} 
As for ScanNet segmentation task, we take SPUNet as the backbone. We leverage Adam as the optimizer and OneCycle as the scheduler. The learning rate and weight decay are set as 3.5e-4 and 0.01, respectively. The batch size is set as 4, and the number of training epochs is 40. Regarding ScanNet detection task, we take FCAF3D as baseline.
As to SemanticKITTI, we follow the training protocol of CENet~\cite{cenet}. The resolution of the range image is chosen as 64$\times$512. Following CENet, we calculate the mIoU on the range image instead of the original point cloud.

\begin{table}[t!]
    \begin{minipage}{.42\textwidth}
        \centering
        \small
        \caption{Performance on ScanNet v2 val set (segmentation).}
        \label{tab:scannet-seg}
        
\begin{tabular}{l | c }
    \toprule Method                                &  mIoU                 \\
    \specialrule{0em}{2pt}{0pt}
    \hline
    \specialrule{0em}{2pt}{0pt}
    PointNet++~\cite{qi2017pointnet++}             &  53.5                 \\
    PointConv~\cite{wu2019pointconv}               &  61.0                 \\
    JointPointBased~\cite{chiang2019unified}       &  69.2                 \\
    PointASNL~\cite{yan2020pointasnl}              &  63.5                 \\
    KPConv~\cite{thomas2019kpconv}                 &  69.2                 \\
    SparseConvNet~\cite{graham20183d}              &  69.3                 \\
    MinkUNet~\cite{choy20194d}                     &  72.2                 \\
    PTv1~\cite{ptv1}                      &  70.6                 \\
    PTv2~\cite{ptv2}                        &  75.4                 \\
    OctFormer~\cite{octformer}                     & 75.7
               \\
    \specialrule{0em}{2pt}{0pt}
    \hline
    \specialrule{0em}{2pt}{0pt}
    \algorithmname~(Ours)                     &  \textbf{77.0}        \\
    \bottomrule
\end{tabular}

    \end{minipage}
    \begin{minipage}{.42\textwidth}
        \centering
        \small
        \vspace{2ex}
        \caption{Performance on ScanNet v2 val set (detection).}
        \label{tab:scannet-det}
        
\begin{tabular}{l | c  c }
    \toprule Method    & mAP$_{25}$   & mAP$_{50}$    \\
    \specialrule{0em}{2pt}{0pt}
    \hline
    \specialrule{0em}{2pt}{0pt}
    VoteNet~\cite{votenet}   & 58.6   &    33.5 \\
    HGNet~\cite{hgnet}      &   61.3    &   34.4  \\
    GSDN~\cite{gsdn}      &    62.8    &     34.8     \\
    3D-MPA~\cite{3d-mpa}      &   64.2   &   49.2    \\
    3DETR~\cite{3detr}       &  65.0  &    47.0 \\
    H3DNet~\cite{h3dnet}     &  67.2  &    48.1 \\
    FCAF3D~\cite{fcaf3d}     & 67.4   &    49.8 \\
    \specialrule{0em}{2pt}{0pt}
    \hline
    \specialrule{0em}{2pt}{0pt}
    \algorithmname~(Ours)    & \textbf{68.8}   & \textbf{51.6}  \\    
    \bottomrule
\end{tabular}

    \end{minipage}
        \begin{minipage}{.42\textwidth}
        \centering
        \small
        \vspace{2ex}
        \caption{Performance on SemanticKITTI val set.}
        \label{tab:semkitti-seg}
        
\begin{tabular}{l | c }
    \toprule Method                                &  mIoU                 \\
    \specialrule{0em}{2pt}{0pt}
    \hline
    \specialrule{0em}{2pt}{0pt}
    CENet~\cite{cenet}      & 65.3 \\
    \specialrule{0em}{2pt}{0pt}
    \hline
    \specialrule{0em}{2pt}{0pt}
    \algorithmname~(Ours)   &  \textbf{66.8}   \\
    \bottomrule
\end{tabular}

    \end{minipage}
\end{table}

\subsection{Quantitative comparison}

We provide quantitative comparison between our \algorithmname~and previous algorithms in ScanNet segmentation, ScanNet detection and SemanticKITTI segmentation benchmarks.

\subsection{ScanNet Segmentation} 
Table~\ref{tab:scannet-seg} summarizes the performance of our \algorithmname~with previous segmentation models, including the Point Transformer series~\cite{ptv1,ptv2} and OctFormer~\cite{octformer}, on the ScanNet segmentation task. It is apparent that our \algorithmname~outperforms previous segmentors by a large margin. Specifically, \algorithmname~surpasses Point Transformer V2 by 1.6 mIoU, which strongly demonstrates the efficacy of the proposed HexPlane representation. 

\subsection{ScanNet Detection} 
FCAF3D~\cite{fcaf3d} is a competitive fully convolutional anchor-free 3D detector and is free from the object geometry assumption. When applying the HexPlane representation as well as the HAM module to FCAF3D on the ScanNet detection task, we can also observe 1.4 mAP$_{25}$ and 1.8 mAP$_{50}$ performance gains, as shown in Table~\ref{tab:scannet-det}. Here, we adopt the 14-layer variant of FCAF3D to accelerate the training process. Our \algorithmname~outperforms previous 3D detectors in terms of mAP$_{25}$ and mAP$_{50}$, including VoteNet~\cite{votenet}, 3DETR~\cite{3detr}, H3DNet~\cite{h3dnet}. The aforementioned results strongly demonstrate the effectiveness of our HexPlane representation.

\subsection{SemanticKITTI Segmentation} 
We apply the HexPlane representation to CENet~\cite{cenet}. CENet is a popular range-based LiDAR segmentor which adopts large kernel convolution, better activation function, and auxiliary prediction loss to enhance the segmentation performance. As presented in Table~\ref{tab:semkitti-seg}, it can boost the performance of CENet from 65.3 mIoU to 66.8 mIoU on SemanticKITTI val set. Compared with the vanilla CENet, our HexPlane representation can not only bring knowledge from other views of the scene, but also benefit from the abundant knowledge from the 2D world. 

\subsection{Limitations}
Despite the good performance in ScanNet and SemanticKITTI benchmarks, there are several shortcomings in the current framework. One potential shortcoming of our method lies in the computation complexity and resource consumption. Since we employ the 2D backbones to extract visual features from the projected 2D views of the point cloud. The additional resource consumption may impede the application of our method to embedded devices with limited computation budget. Another limitation is that our work mainly concentrates on semantic segmentation and the efficacy of our algorithm has not been verified on various indoor and outdoor point cloud detection tasks. In the future, we will also explore the possibility of designing learnable projection module to adaptively project 3D point clouds to the 2D plane.

\section{CONCLUSION}
\label{sec:conclusion}
In this paper, we design the HexPlane representation for 3D perception tasks. Compared to the point and voxel representations, the HexPlane representation is much more efficient and can utilize the highly optimized 2D operations to process the point clouds. It can also leverage off-the-shelf 2D models, network weights as well as the training recipes to fulfill precise 3D scene understanding. The HexPlane representation can inject the abundent 2D knowledge into the 3D world, easily utilize the rich knowledge of the off-the-shelf 2D models, and relieve the data scarcity problem in the 3D field. VPM and HAM are specifically devised to achieve view projection and HexPlane feature fusion, respectively. VPM can maximally retain the original spatial information and HAM can adaptively fuse the most informative information for each point. The proposed \algorithmname~exhibits compelling performance on ScanNet LiDAR segmentation and detection tasks as well as SemanticKITTI LiDAR segmentation task. Notably, on the ScanNet 3D segmentation task, our method obtains \textbf{77.0} mIoU on the validation set, surpassing Point Transformer V2 by \textbf{1.6} mIoU. Our method can be seamlessly incorporated into contemporary voxel-based, point-based and range-based algorithms, and bring considerable gains without bells and whistles. We hope our \algorithmname~can serve as the footstone of the 3D semantic scene understanding field. 

\bibliographystyle{IEEEtran}
\bibliography{IEEEabrv, triplane}

\end{document}